\documentclass[12pt]{article}
\usepackage[utf8]{inputenc}
\usepackage{hyperref}
\usepackage{graphicx}
\usepackage{amsmath}
\usepackage{mathptmx} % Times New Roman font
\usepackage{geometry} % margins
\usepackage{authblk} % affiliations
\usepackage{indentfirst} % indent first paragraph

\makeatletter
\def\@maketitle{%
  \newpage
  \begin{center}%
  \let \footnote \thanks
    {\large\bfseries \@title \par}%
    \vskip \@affilsep%
    {\large%
        \begin{tabular}[t]{c}%
        \@author\\
        contact: \href{mailto:\contact}{\contact}
      \end{tabular}\par}%
  \end{center}}
\makeatother
\title{A Few-Shot Learning Approach for Accelerated MRI via Fusion of Data-Driven and Subject-Driven Priors}
\def\contact{salman@ee.bilkent.edu.tr}
\author[1,2]{Salman Ul Hassan Dar}
\author[1,2]{Mahmut Yurt}
\author[1,2,3]{Tolga \c{C}ukur}

\affil[1]{Department of Electrical and Electronics Engineering, Bilkent University, Ankara, Turkey}
\affil[2]{National Magnetic Resonance Research Center (UMRAM), Bilkent University, Ankara, Turkey}
\affil[3]{Neuroscience Program, Aysel Sabuncu Brain Research Center, Bilkent University, Ankara, Turkey}

\begin{document}
\maketitle
\section*{\large Synopsis}
Deep neural networks (DNNs) have recently found emerging use in accelerated MRI reconstruction. DNNs typically learn data-driven priors from large datasets constituting pairs of undersampled and fully-sampled acquisitions. Acquiring such large datasets, however, might be impractical. To mitigate this limitation, we propose a few-shot learning approach for accelerated MRI that merges subject-driven priors obtained via physical signal models with data-driven priors obtained from a few training samples. Demonstrations on brain MR images from the NYU fastMRI dataset indicate that the proposed approach requires just a few samples to outperform traditional parallel imaging and DNN algorithms.
\section*{\large Introduction}
A mainstream framework for reconstruction of accelerated MR acquisitions rests on deep neural network (DNN) architectures\textsuperscript{1-11}. To recover images given undersampled acquisitions, DNNs typically learn data-driven priors from large training datasets in a supervised fashion\textsuperscript{1-10}. While DNNs have shown remarkable performance, compilation of large-scale datasets for each anatomy and each protocol is challenging. To mitigate this issue, here we propose a few-shot learning approach for accelerated MRI. The proposed approach consists of a composite deep neural network (COMNET) that fuses subject-driven priors obtained via a physical signal model with data-driven priors obtained from only few training samples.
\begin{figure}
    \centering
    \includegraphics[width=1\textwidth]{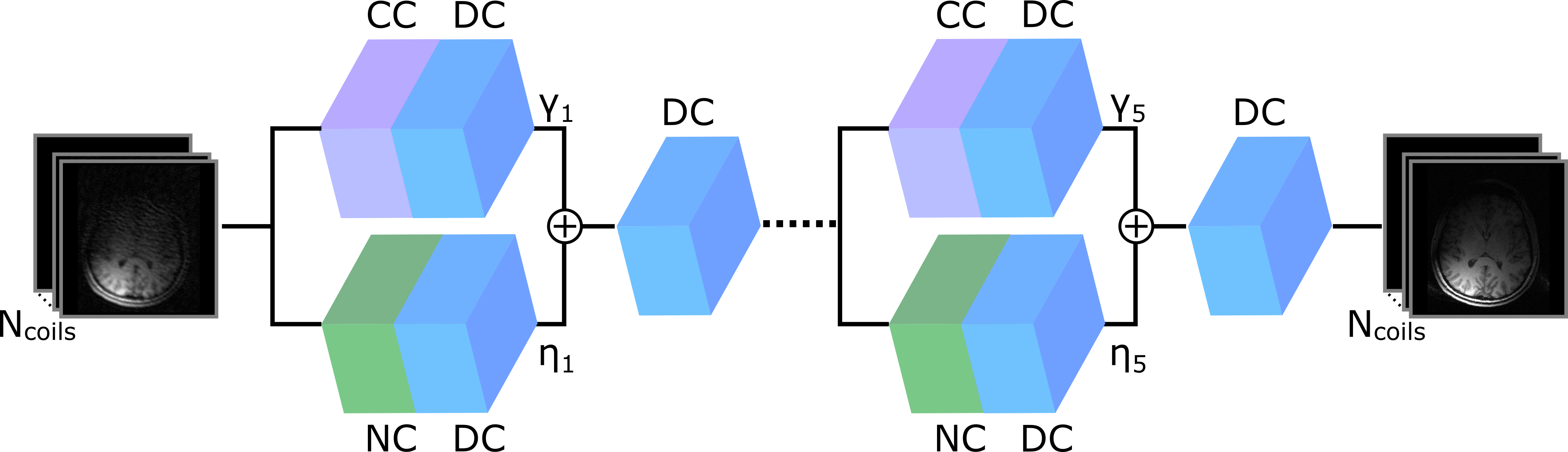}
    \caption{COMNET consists of an unrolled cascade of sub-networks where each sub-network consists of a calibration consistency (CC) block fused with a network consistency (NC) block, both followed by a data consistency (DC) block.}
    \label{fig:fig1}
\end{figure}
\section*{\large Methods}
The reconstruction problem in COMNET can be formulated as: 
\begin{align}
    \widehat{x} = \underset{x}{\arg\min} \quad \lambda \underbrace{||F_ux-y||_2}_\text{Data consistency}+ \underbrace{(G-I)Fx}_\text{Subject-driven Prior}+ \underbrace{|| C(A^*x^u;\theta^*) - A^*x||_2}_\text{Data-driven Prior}
\end{align}
where $F_u$ is the partial Fourier operator defined at the sampled k-space locations, $x$ is the image to be reconstructed, $y$ are the acquired k-space data, $G$ is a linear operator enforcing consistency with a fully-sampled auto-calibration region, $C$ denotes the purely learning-based model for reconstruction, $x^u$ is the Fourier reconstruction of undersampled data, and $A$ and $A^*$ denote coil sensitivity maps and their conjugate obtained via ERPIRiT\textsuperscript{12}. COMNET comprises three blocks to enforce data consistency (DC), to enforce subject-driven priors via calibration consistency (CC), and to enforce data-driven priors via network consistency (NC) terms in the objective. A common approach is to connect these blocks in series in an unfolded architecture, and solve the optimization problem by alternating minimization of individual terms\textsuperscript{10}. However, this serial structure introduces undesirable dependency among consecutive blocks that can lead to information loss. To address this problem, here we proposed to fuse information from parallel connected NC and CC blocks (Figure 1). An unrolled cascade of subnetworks are then leveraged for image recovery, and the output of the $pth$ subnetwork receiving input from the previous subnetwork is given by:
\begin{align}
x_p=f_{DC}(A\gamma_p A^*f_{DC}(f_{NC}(A^*x_{p-1}))+A\eta_p A^*f_{DC}(f_{CC}(x_{p-1})))
\end{align}
where $f_{CC}$, $f_{NC}$, and $f_{DC}$ denote mappings by CC, NC and DC blocks, $x_p$ is the output of the $pth$ sub-network, $x_{p-1}$ is the output of the $(p-1)th$ sub-network, and $\gamma_p$ and $\eta_p$ are fusion parameters to combine information from the NC and CC blocks. NC block was adopted from \textsuperscript{2} where each network consisted of 1 input layer, 4 convolutional layers each containing 64 channels, and 1 output layer. Real and imaginary parts were recovered using separate network branches. The CC block was implemented via SPIRiT\textsuperscript{13} where 5 CC projections were performed within each block. The network was trained in an end-to-end manner, where parameters of each sub-network was identical except for the weighing parameters ($\gamma$ and $\eta$) that were different for each sub-network. ADAM optimizer\textsuperscript{14} was used with a learning rate of $10^{-4}$, and parameters $\beta_1$=0.90 and $\beta_2$=0.99. Network was trained to minimize $\ell_1$ and $\ell_2$ norm difference between reconstructed and ground-truth images. Number of epochs was set to 200. 
Demonstrations were performed on contrast enhanced T1-weighted (cT1), T2-weighted and FLAIR images from the NYU fastMRI dataset\textsuperscript{15}. 30 subjects were reserved for training, 10 for validation and 40 for testing. For a systematic evaluation, cT1 and FLAIR images were cropped to a final size of 256x320x10 and T2 images were cropped to 288x384x10 when necessary. Geometric coil compression\textsuperscript{16} was utilized to ensure that all MRI data had 5 coils. Acquisitions were retrospectively undersampled at R=4x via random sampling masks generated using normal sampling density. COMNET was compared against a regular DNN consisting of only data-driven priors, and L1-SPIRiT consisting of subject-driven priors coupled with sparsity prior in the Wavelet domain. The number of samples for COMNET and DNN were varied from 2 to 300. All hyperparameters were selected via cross-validation with three-way split of data.
\section*{\large Results}
Average PSNR and SSIM values of recovered cT1- weighted, T2-weighted and FLAIR images at R=4x are listed in Table. 1. Both DNN and COMNET were trained on 6 cross-sections from a single subject. On average, COMNET achieves 0.48dB higher PSNR and 0.72\% higher SSIM compared to the second-best method. 
Figure 2 shows PSNR values across recovered cT1- weighted, T2-weighted and FLAIR images as a function of number of training samples. DNN, on average, requires around 90 cross-sections from 9 subjects to outperform L1-SPIRiT. COMNET, on the other hand, requires 2,4, and 6 cross-sections from just a single subject to outperform L1-SPIRiT on cT1- weighted, T2-weighted and FLAIR images. Importantly, COMNET reduces the number of required samples by at least an order of magnitude compared to DNN.
Figures 3 and 4 show representative T2-weighted and FLAIR images from L1-SPIRiT, DNN and COMNET. DNN and COMNET were trained on 6 cross-sections from a single subject. COMNET outperforms both L1-SPIRiT and DNN in terms of residual aliasing artifacts.
\section*{\large Discussion}
Here, we propose a few-shot learning approach for MR image reconstructions using deep neural networks. The proposed approach synergistically combines subject-driven priors with data-driven priors to address the issue of data scarcity in DNNs for MR image reconstruction.
\section*{\large Conclusion}
The proposed approach enables data-efficient training of deep neural networks for MR image reconstruction. Therefore, COMNET holds great promise for improving practical use of deep learning models in accelerated MRI.
\section*{\large Acknowledgements}
This work was supported in part by a TUBA GEBIP fellowship, by a TUBITAK 1001 Grant (118E256), and by a BAGEP fellowship awarded to T. Çukur. We also gratefully acknowledge the support of NVIDIA Corporation with the donation of the Titan X Pascal GPU used for this research.
\section*{\large References}
1. Hammernik K, Klatzer T, Kobler E, et al. Learning a Variational Network for Reconstruction of Accelerated MRI Data. Magn. Reson. Med. 2017;79:3055–3071.

2. Schlemper J, Caballero J, Hajnal J V., Price A, Rueckert D. A Deep Cascade of Convolutional Neural Networks for MR Image Reconstruction. In: International Conference on Information Processing in Medical Imaging. ; 2017. pp. 647–658.

3. Mardani M, Gong E, Cheng JY, et al. Deep Generative Adversarial Neural Networks for Compressive Sensing (GANCS) MRI. IEEE Trans. Med. Imaging 2018;38:167-179. %doi: 10.1109/TMI.2018.2858752.

4. Han Y, Yoo J, Kim HH, Shin HJ, Sung K, Ye JC. Deep learning with domain adaptation for accelerated projection-reconstruction MR. Magn. Reson. Med. 2018;80:1189–1205. %doi: 10.1002/mrm.27106.

5. Wang S, Su Z, Ying L, et al. Accelerating magnetic resonance imaging via deep learning. In: IEEE 13th International Symposium on Biomedical Imaging (ISBI). ; 2016. pp. 514–517. %doi: 10.1109/ISBI.2016.7493320.

6. Yu S, Dong H, Yang G, et al. DAGAN: Deep de-aliasing generative adversarial networks for fast compressed sensing MRI reconstruction. IEEE Trans. Med. Imaging 2018;37:1310–1321.

7. Zhu B, Liu JZ, Rosen BR, Rosen MS. Image reconstruction by domain transform manifold learning. Nature 2018;555:487–492. %doi: 10.1017/CCOL052182303X.002.

8. Aggarwal HK, Mani MP , Jacob M. MoDL: Model-Based Deep Learning Architecture for Inverse Problems. IEEE Trans. Med. Imaging 2018;38:394-405. %doi: 10.1017/CCOL052182303X.002.

9. Eo T, Jun Y, Kim T, Jang J, Lee H, Hwang D. KIKI‐net: cross‐domain convolutional neural networks for reconstructing undersampled magnetic resonance images. Magn. Reson. Med. 2018;80:2188–2201.

10. Dar SUH, \"{O}zbey M, \c{C}atl{\i} AB, \c{C}ukur T. A Transfer-Learning Approach for Accelerated MRI Using Deep Neural Networks. Magn. Reson. Med. 2020;84:663–685. %doi: 10.1002/mrm.28148.

11. Ak\c{c}akaya M, Moeller S, Weing\"{a}rtner S, U\u{g}urbil K. Scan-specific robust artificial-neural-networks for k-space interpolation (RAKI) reconstruction: Database-free deep learning for fast imaging. Magn. Reson. Med. 2019;81:439–453. %doi: 10.1002/mrm.27420.

12. Uecker M, Lai P, Murphy MJ, et al. ESPIRiT-an eigenvalue approach to autocalibrating parallel MRI: Where SENSE meets GRAPPA. Magn. Reson. Med. 2014;71:990–1001. %doi: 10.1002/mrm.24751.

13. Lustig M, Pauly JM. SPIRiT: Iterative self-consistent parallel imaging reconstruction from arbitrary k-space. Magn. Reson. Med. 2010;64:457–71. %doi: 10.1002/mrm.22428.

14. Kingma DP, Ba JL. Adam: a Method for Stochastic Optimization. In: International Conference on Learning Representations. 2015. %doi: http://doi.acm.org.ezproxy.lib.ucf.edu/10.1145/1830483.1830503.

15. Zbontar J, Knoll F, Sriram A, et al. fastMRI: An open dataset and benchmarks for accelerated MRI. arXiv 2018.

16. Zhang T, Pauly JM, Vasanawala SS, Lustig M. Coil compression for accelerated imaging with Cartesian sampling. Magn. Reson. Med. 2013;69:571–82. %doi: 10.1002/mrm.24267. 

\newpage
\begin{figure}[hbt!]
    \centering
    \includegraphics[width=0.7\textwidth]{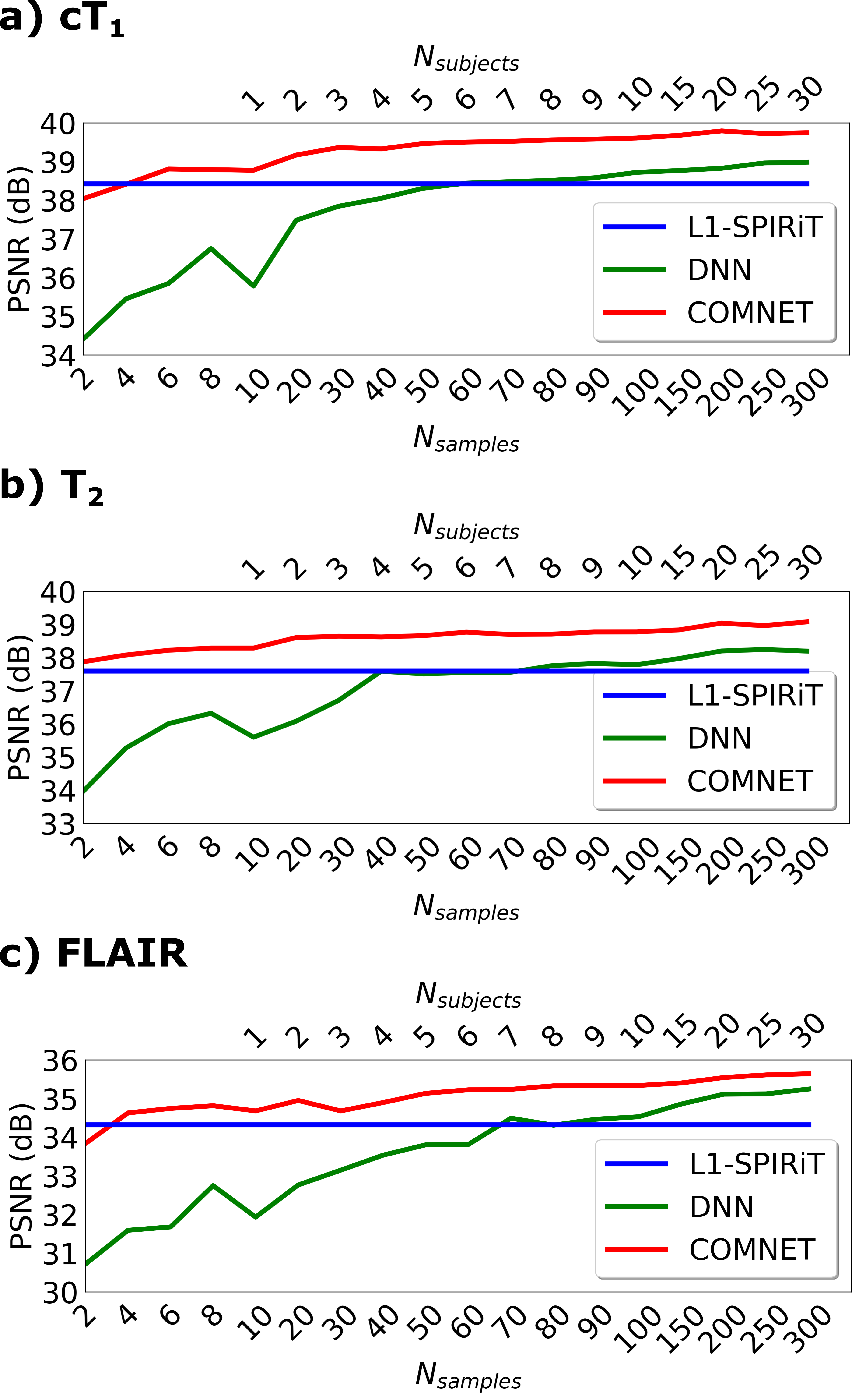}
    \caption{Average PSNR values of a) cT1, b) T2, and c) FLAIR images of test subjects as a function of number of training subjects (upper x-axis), and training samples (lower x-axis). COMNET requires just a few training samples from a single subject to outperform L1-SPIRiT. On the other hand, DNN on average requires around 90 samples from 9 different subjects to start performing better than L1-SPIRiT.}
    \label{fig:plots}
\end{figure}
\begin{figure}[hbt!]
    \centering
    \includegraphics[width=0.9\textwidth]{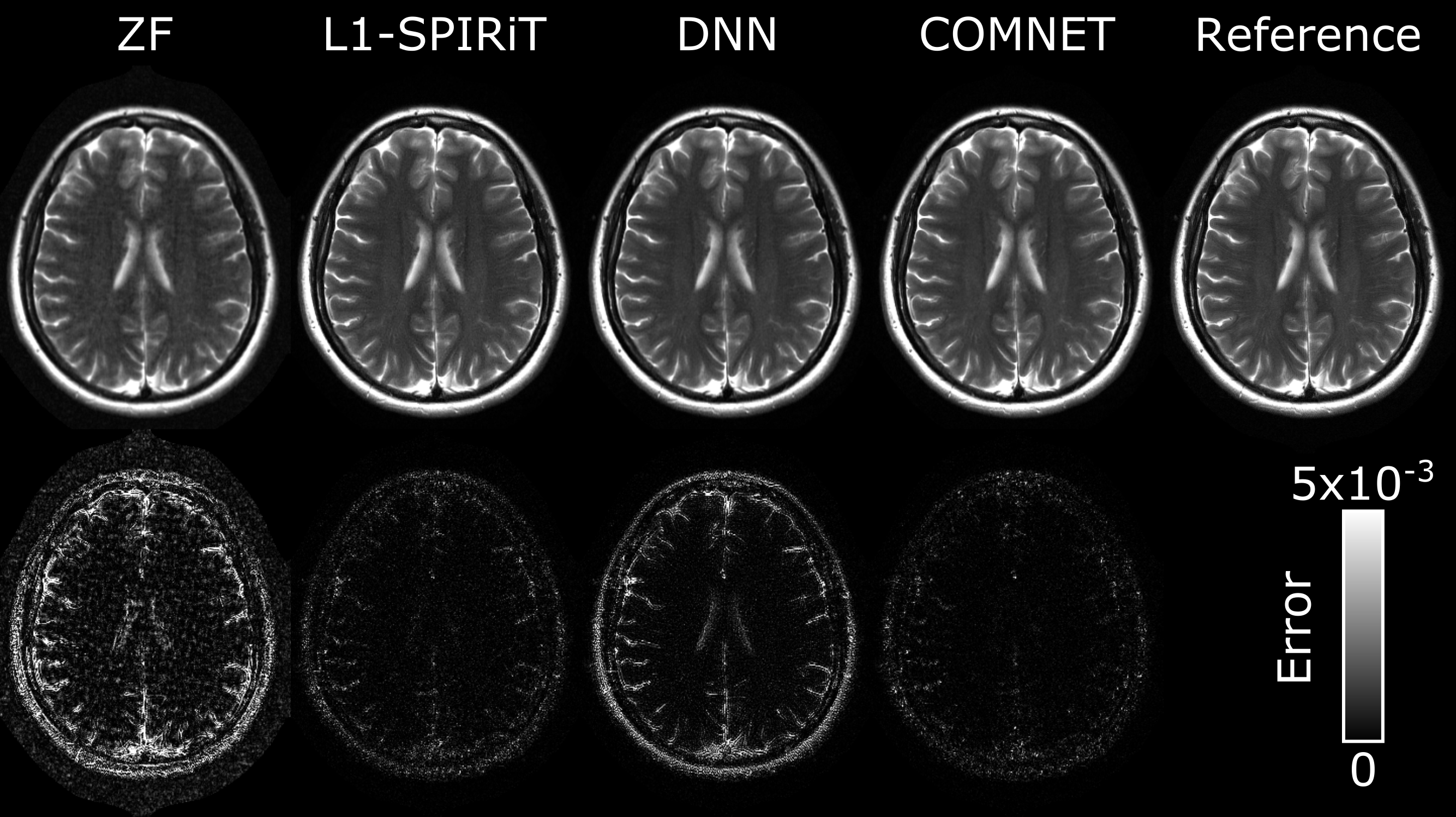}
    \caption{Recovered T2-weighted images via L1-SPIRiT, DNN, and COMNET are shown along with the corresponding squared error maps with the zero-filled (ZF) reconstruction and reference image. DNN and COMNET were trained on 6 cross-sections from a single subject. COMNET shows superior performance to DNN and L1-SPIRiT in terms of residual aliasing artifacts.}
    \label{fig:t2_slices}
\end{figure}

\begin{figure}[hbt!]
    \centering
    \includegraphics[width=0.9\textwidth]{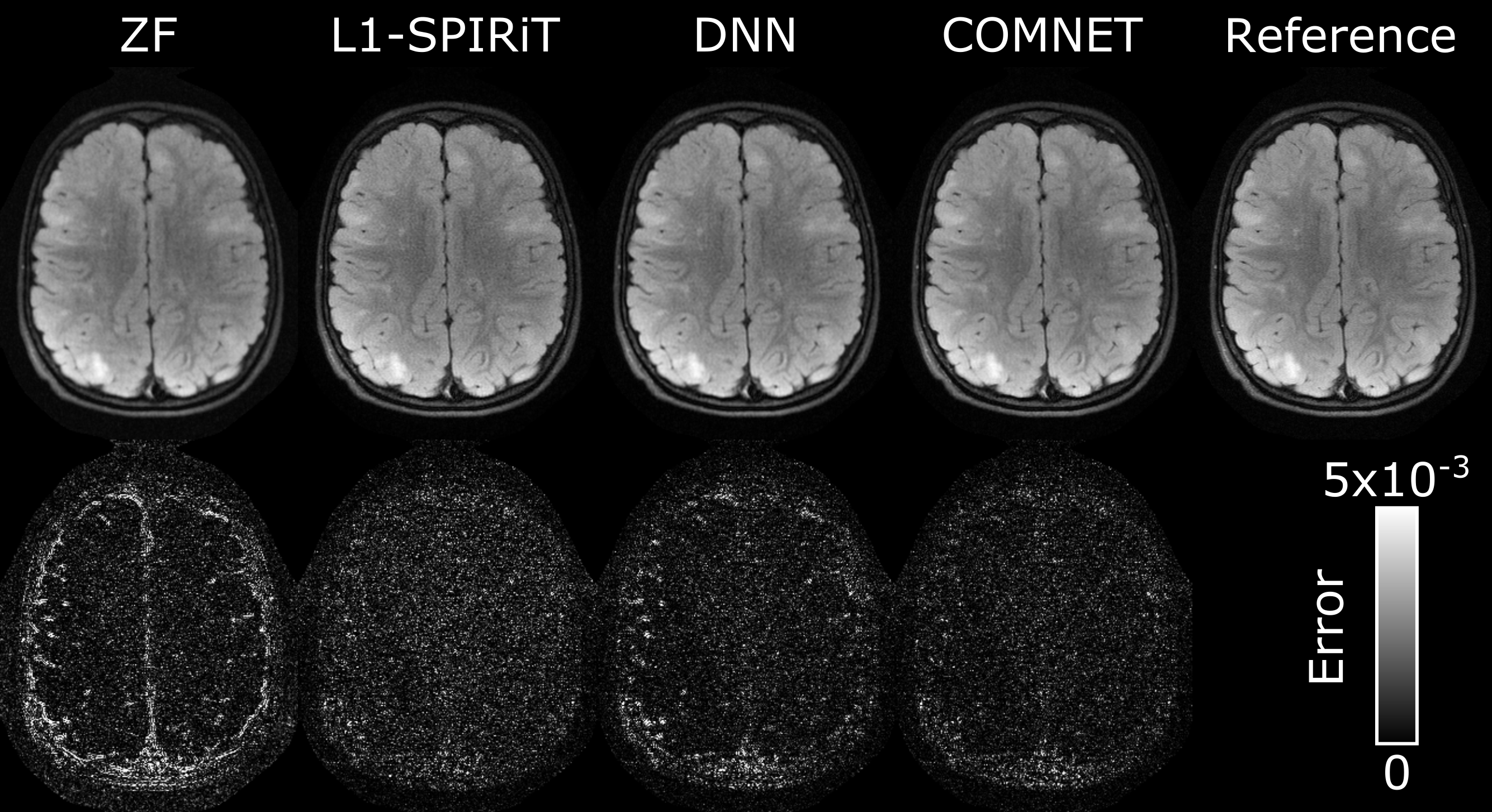}
    \caption{Recovered FLAIR images via L1-SPIRiT, DNN, and COMNET are shown along with the corresponding squared error maps with the zero-filled (ZF) reconstruction and reference image. DNN and COMNET were trained on 6 cross-sections from a single subject. COMNET shows superior performance to DNN and L1-SPIRiT in terms of residual aliasing artifacts.}
    \label{fig:flair_slices}
\end{figure}

\newpage
\begin{table}[h]
    \centering
    \begin{tabular}{|c|c|c|c|c|c|c|}
    \hline
         &\multicolumn{2}{|c|}{L1-SPIRiT} & \multicolumn{2}{|c|}{DNN}& \multicolumn{2}{|c|}{COMNET} \\
         \hline
         & PSNR & SSIM & PSNR & SSIM & PSNR & SSIM \\
         \hline
         cT1 &  38.43$\pm$0.24 & 94.65$\pm$0.19 & 35.85$\pm$0.24 & 94.30$\pm$0.17 & \textbf{38.80}$\pm$\textbf{0.24} & \textbf{95.54}$\pm$\textbf{0.16} \\
         \hline
         T2 &  37.60$\pm$0.15 & 95.60$\pm$0.09 & 36.01$\pm$0.18 & 95.74$\pm$0.08 & \textbf{38.22}$\pm$\textbf{0.16} & \textbf{96.51}$\pm$\textbf{0.07} \\
         \hline
         FLAIR &  34.32$\pm$0.44 & 90.31$\pm$1.03 & 31.68$\pm$0.46 & 91.13$\pm$0.87 & \textbf{34.75}$\pm$\textbf{0.43} & \textbf{91.63}$\pm$\textbf{0.97} \\
         \hline
    \end{tabular}
    \caption{PSNR (dB) and SSIM (\%) values of recovered cT1-weighted, T2-weighted, and FLAIR images. DNN and COMNET were trained on 6 samples from a single subject. Best performing models are marked with bold font.}
    \label{tab:table}
\end{table}
\end{document}